\documentclass[conference]{IEEEtran}
\IEEEoverridecommandlockouts
\usepackage{cite}
\usepackage{amsmath,amssymb,amsfonts,bm}
\usepackage{amsthm}

\usepackage{algorithm}
\usepackage{algorithmic}
\usepackage{dsfont}
\usepackage{graphicx}
\usepackage{textcomp}
\usepackage{xcolor}
\usepackage{siunitx}
\usepackage[hidelinks]{hyperref}
\usepackage{booktabs}
\usepackage{todonotes}
\usepackage{multirow}

\def\BibTeX{{\rm B\kern-.05em{\sc i\kern-.025em b}\kern-.08em
    T\kern-.1667em\lower.7ex\hbox{E}\kern-.125emX}}

\newcommand{\Kmppi}{\ensuremath{K}}
\newcommand{\lambdamppi}{\ensuremath{\lambda}}

\newcommand{\SO}{\mathbb{S}\mathbb{O}}
\newcommand{\real}{\mathbb{R}}
\newcommand{\x}{\bm{x}}

\renewcommand{\u}{\bm{u}}
\newcommand{\pos}{\bm{p}}
\newcommand{\rot}{\bm{q}}
\newcommand{\vel}{\bm{v}}
\newcommand{\angvel}{\bm{\omega}}
\newcommand{\dt}{\Delta t}

\renewcommand{\S}{\mathcal{S}}

\newcommand{\ub}{\bm{u}}    

\begin{document}

\title{OA-MPPI: Occlusion-Aware Model Predictive Path Integral Control for UAV Flight}

\author{\IEEEauthorblockN{Vittorio Palladino, Teaya Yang, Ruiqi Zhang, Mark W.~Mueller}
\thanks{The authors are with the High Performance Robotics Laboratory, Department
of Mechanical Engineering, University of California, Berkeley, CA 94720,
United States. Email: \{vpall, teaya.yang, richzhang, mwm\}@berkeley.edu}}

\maketitle

\begin{abstract}
Autonomous UAV flight through cluttered and partially unknown environments requires reasoning not only about observed obstacles but also about occluded regions that the sensor cannot observe. We present OA-MPPI, an obstacle- and occlusion-aware extension of Model Predictive Path Integral (MPPI) control for quadrotor flight that accounts for potential moving agents emerging from these regions into the vehicle’s path. At every planning step, we extract a 3D occlusion boundary from the online occupancy map and use it to model the regions that hidden agents could reach over the prediction horizon. We penalize trajectories that enter these expanding regions within MPPI rollouts generated using nonlinear quadrotor dynamics and accounting for individual rotor thrust limits. We validate the proposed approach in simulation and hardware flight experiments, with the complete pipeline running onboard the vehicle in real time. Results show increased clearance from occlusion boundaries compared to baseline MPPI in both settings, as well as avoidance of an agent emerging from occlusion in simulation.
\end{abstract}

\noindent\textbf{Supplementary Video:}\ \url{https://youtu.be/oSeQau1mG1g}\\

\section{Introduction}
Control of Uncrewed Aerial Vehicles (UAVs) in cluttered and unknown environments is a challenging problem, and one that is becoming increasingly critical as UAVs are deployed for search-and-rescue missions~\cite{lyu2023rescue},
infrastructure inspection~\cite{Tordesillas2019FASTER}, and aerial coverage in
disaster response~\cite{recchiuto2018disaster}. These applications impose three
coupled challenges: (1) the controller must efficiently account for nonlinear vehicle dynamics and actuator limits; (2) it must detect and avoid obstacles using onboard sensing in real time; (3) it must anticipate potential collisions with moving obstacles emerging from unobserved regions.

The first two challenges are traditionally addressed with Nonlinear Model Predictive Control (NMPC), which incorporates vehicle dynamics, actuator limits, and collision-avoidance constraints within a unified optimization framework. However, the resulting nonconvex optimization problems can be computationally demanding to solve in real time, particularly in cluttered environments. Model Predictive Path Integral (MPPI) control~\cite{williams2017originMPPI, Minark2024ModelPP} offers a sampling-based alternative that evaluates candidate trajectories through simulated rollouts without requiring derivatives of the dynamics or cost function. This enables the direct evaluation of complex, nonconvex collision costs and provides a flexible foundation for incorporating the risks posed by unobserved regions.

\begin{figure}[!t]
  \centering
  \includegraphics[width=\columnwidth]{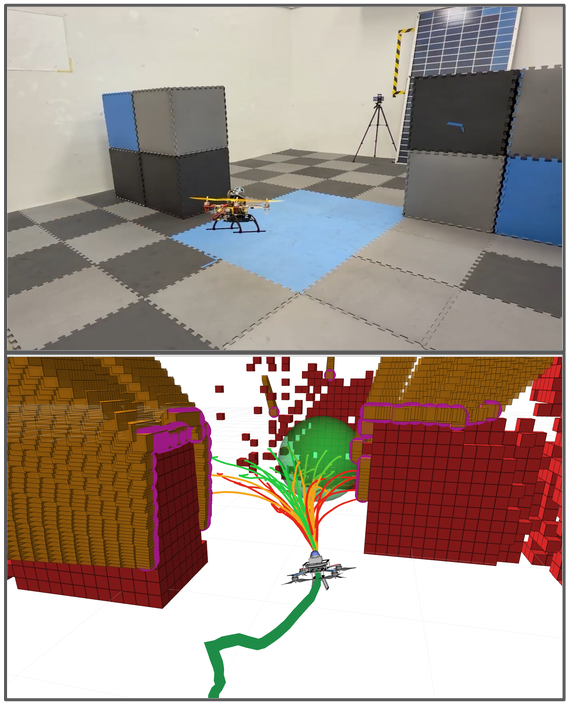}
  \caption{Example of OA-MPPI in operation. Top: experimental setup. Bottom: occupancy map and sampled rollouts. Red voxels denote occupied space, and orange voxels mark the extracted occlusion boundary. Green lines indicate collision-free rollouts, yellow lines indicate rollouts entering an expanding keep-out region, and red lines indicate rollouts colliding with mapped obstacles. The green sphere marks the goal. Ground voxels
are omitted in the plot for clarity.}
  \label{fig:first_page_example}
\end{figure}

The third challenge has received comparatively less attention in aerial robotics and has instead
been studied mainly in the context of autonomous ground vehicles, where
occlusions from parked cars or buildings can hide pedestrians or oncoming
traffic at intersections. There, occlusion-aware planners typically detect
the occluded region directly from the range sensor and reason about
phantom agents that could be hiding within
it~\cite{Firoozi2022OAMPCOM, Park2023OcclusionAwareRA}, bounding the risk they
pose through a forward-reachability or risk-assessment
formulation before the vehicle commits to a trajectory.

In this paper, we tackle all three challenges jointly by building a unified MPPI-based control framework for UAV flight. The framework identifies the 3D boundaries of regions occluded by observed obstacles and expands a potential collision region around these boundaries over the prediction horizon to account for agents that may emerge from occluded regions. The resulting occlusion-aware cost is evaluated alongside an observed-obstacle avoidance cost within the MPPI rollouts. We evaluate the approach in simulation and real-time hardware experiments, using a LiDAR sensor for both state estimation and obstacle detection. Our results demonstrate that accounting for occluded regions produces safer trajectories near occlusion boundaries than a controller without the occlusion-aware cost.

\begin{figure*}[!t]
  \centering
  \includegraphics[width=\textwidth]{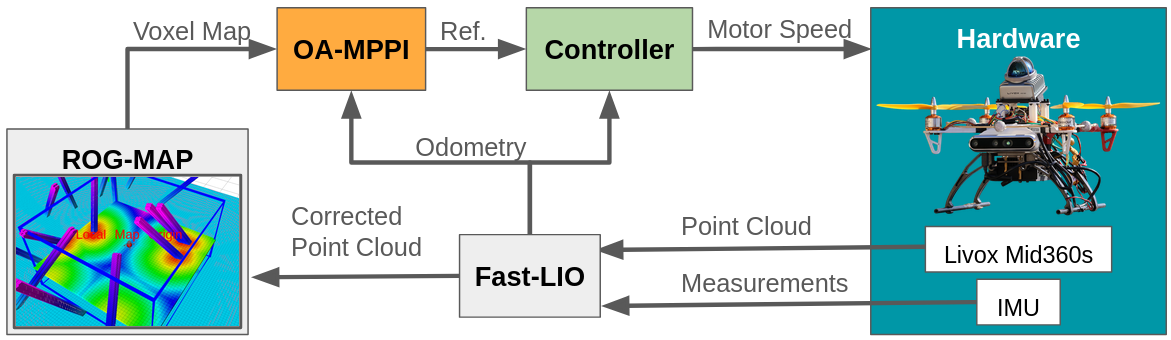}
  \caption{Overview of the OA-MPPI framework. ``Ref.'' denotes the reference trajectory sent to the controller.}
  \label{fig:occlusion_boundaries_first_page}
\end{figure*}

\section{Related Work} \label{section:related_work}

\subsubsection*{Occlusion-aware planning}
Reasoning about occluded dynamic agents has mostly been studied for
ground vehicles. Early work accounts for occlusion by restricting planning to the visible region~\cite{Chung2009SafeNav} or avoiding states from which the vehicle cannot stop before a collision ~\cite{Bouraine2014PassivelySafe}. Other approaches explicitly model phantom agents in occluded regions and account for their possible motion during planning. These methods differ in how they identify occluded regions, predict hidden-agent motion, and incorporate the resulting risk. Firoozi et al.~\cite{Firoozi2022OAMPCOM} detect occlusion boundaries from LiDAR range discontinuities and constrain an NMPC planner using expanding reachable sets. Park et al.~\cite{Park2023OcclusionAwareRA} quantify hidden-agent reachability using simplified state distributions and impose risk-dependent speed limits. Koschi and Althoff~\cite{Koschi2020SetBased} use known road geometry and traffic rules to conservatively predict the occupancy of hidden traffic participants. Möller et al.~\cite{Moller2025Shadows} track occluded regions over time and evaluate candidate trajectories using agent-specific motion predictions and collision risk. We extend OA-MPC's worst-case hidden-agent approach~\cite{Firoozi2022OAMPCOM} to 3D flight by incorporating a collision-risk cost into an MPPI controller that accounts for nonlinear quadrotor dynamics. Our occlusion detection uses occupancy maps, removing OA-MPC's dependence on repeated LiDAR scan patterns and supporting any sensor capable of generating the required occupancy map.

\subsubsection*{MPPI for aerial robots}

Unlike gradient-based trajectory optimization methods, which require a
smooth cost function with well-defined gradients everywhere, MPPI~\cite{williams2017originMPPI}
only evaluates costs forward along sampled rollouts, without ever
differentiating them. This allows MPPI to directly consume costs derived
from raw occupancy grids or distance fields, including the discontinuities
and gradient singularities that arise near non-convex obstacle boundaries,
without requiring a smoothed or convex approximation. This flexibility
makes MPPI well suited to jointly evaluating obstacle and occlusion costs
within rollouts of nonlinear quadrotor dynamics. Minarik et al.~\cite{Minark2024ModelPP} demonstrate onboard MPPI control for quadrotors, tracking references at speeds up to \SI{44}{\kilo\meter\per\hour}. Their controller evaluates rollouts in parallel on a GPU and incorporates obstacle avoidance through a binary collision cost. PA-MPPI~\cite{Zhai_2026} adds a perception term that rewards trajectories observing unknown regions when the goal is occluded. However, neither method explicitly models agents emerging from occluded regions. Our cost accounts for the collision risk posed by these agents even when the planned trajectory remains entirely within known free space.

\subsubsection*{Planning in unknown environments}
Several existing aerial planning methods explicitly distinguish known free space from unexplored regions. FASTER~\cite{Tordesillas2019FASTER} plans through both while maintaining a safe backup trajectory within known free space. SUPER~\cite{ren2025safety} similarly combines exploratory planning with a backup trajectory. HDSM~\cite{Toumieh2024HighSpeed} accounts for unexplored space in decentralized swarm planning while avoiding obstacles and other agents. RAPTOR~\cite{zhou2021raptor} refines the flight trajectory and plans yaw to observe potentially hazardous unknown obstacles early enough for avoidance. These methods address navigation through partially mapped environments under limited sensing. Our approach additionally predicts how agents hidden in unknown space could occupy currently known free space over the planning horizon and penalizes trajectories that intersect their potential future occupancy.
\section{Preliminaries} \label{section:preliminaries}

% \subsection{Notation} \label{sec:notation}

% Vectors are written in bold. Throughout the paper the lower index $j$ denotes
% the stage within the prediction horizon and the upper index $k$ the rollout,
% so that $\pos^k_j$ is the position predicted at stage $j$ of rollout $k$.
% The vehicle state is $\x = (\pos, \rot, \vel, \angvel) \in \real^3 \times \SO(3)
% \times \real^3 \times \real^3$, collecting position, attitude, linear velocity
% and body rates, and the control input is $\u = (F, \angvel_{\text{des}})$,
% the collective thrust and the desired body rates. The horizon length is
% $\Tmppi$, the number of rollouts $\Kmppi$, the sampling covariance $\Sigma$,
% the temperature $\lambdamppi$ and the discretization step $\dt$; $\S_k$ is the
% cost of rollout $k$ and $\omega_k$ its importance weight. We write
% $\mathds{1}[\cdot]$ for the indicator function, $\|\cdot\|_2$ for the Euclidean
% norm and $\|\bm{y}\|^2_{P} = \bm{y}^\top P \bm{y}$ for a weighted quadratic
% form. On the perception side, $\mathcal{M}$ denotes the occupancy map,
% $\mathcal{M}_{\text{inf}}$ its inflated layer, $r_m$ the map resolution and
% $r_{\text{veh}}$ the vehicle radius; $\mathcal{B}$ is the occlusion boundary
% and $d_{\mathcal{B}}$ the associated distance field. Quantities carrying a hat,
% such as $\hat{\x}$, are estimated online.
% \subsection{Model Predictive Path Integral Control}
\subsection{Model Predictive Path Integral Control}
\label{subsection:mppi}

MPPI draws a large number of candidate action sequences via Monte
Carlo sampling and combines them into an improved sequence based on their
cost. For a single rollout $k$, the cost takes the general receding-horizon
form
\begin{equation}
  \S_k = V(\x_H) + \sum_{j=0}^{H-1} \ell(\x_j, \u_j), \quad \x_0 = \hat{\x}, \; \x_{j+1} = f(\x_j, \u_j), \label{eq:generic_mpc}
\end{equation}
where $V(\x_H)$ is the terminal cost, $\ell(\x_j, \u_j)$ the stage cost
applied at every step of the horizon $H$, and $\x_j$ the state of rollout $k$
at stage $j$. Each of the $\Kmppi$ rollouts is generated by perturbing the
nominal action sequence from the previous iteration with Gaussian noise
$\delta \u_j^k \sim \mathcal{N}(0, \Sigma)$, all initialized at
$\bm{x}_0^k = \hat{\x}$ and propagated with ${\bm{f}}$, the
quadrotor dynamics of \autoref{sec:quadrotor_dynamics}.

Each rollout is evaluated with a task-specific, possibly non-convex and
non-differentiable, cost function, and the resulting costs
$(\S_1, \dots, \S_{\Kmppi})$ are converted into normalized importance
weights through a softmax transform,
\begin{equation}
  \begin{aligned}
    \omega_k & = \frac{1}{\eta}\exp{\left( -\frac{1}{\lambdamppi} \left(\S_k - \rho\right)\right)},   \\
    \eta     & = \sum_{k=1}^{\Kmppi} \exp{\left( -\frac{1}{\lambdamppi} \left(\S_k - \rho\right)\right)}, \\
    \rho     & = \min \{\S_1, \dots, \S_{\Kmppi}\},
  \end{aligned}\label{eq:cost_2}
\end{equation}
where the minimum cost $\rho$ is subtracted for numerical stability and
$\lambdamppi > 0$ is a temperature parameter trading off averaging all
disturbances ($\lambdamppi \to \infty$) against greedily selecting the best
rollout ($\lambdamppi \to 0$). The nominal sequence is then updated using a softmax function of the sampled disturbances at every stage $j$,
and only the first action $\bm{u}^{\text{nom}}_0$ of the updated sequence is
applied to the system; the remaining actions become the nominal sequence for
the next planning cycle, and the horizon shifts forward by one step.

\begin{table}[htbp]
  \small
  \centering
  \caption{MPPI horizon/sampling parameters and cost weights used.}
  \vspace{-0.3em}
  \addtolength{\tabcolsep}{-0.1em}
  \begin{tabular}{l r | l r}
    \toprule
    \multicolumn{2}{c|}{MPPI param.} & \multicolumn{2}{c}{Cost definition constants} \\
    \midrule
    $H$           & $30$              & $w_{\text{goal}}$      & $0.1$ \\
    $\Kmppi$      & $500$             & $w_{\text{goal}}^{\text{term}}$ & $5.0$ \\
    $\lambdamppi$ & $0.1$             & $w_{\text{col}}$       & $50.0$ \\
    $\dt$         & $\SI{0.1}{\second}$ & $w_{\text{vel}}$     & $15.0$ \\
    \midrule
    \multicolumn{2}{c|}{}             & $\Sigma$ & $\text{diag}(0.60, 0.15, 0.15, 0.05)$ \\
    \multicolumn{2}{c|}{}             & $R$      & $\text{diag}(0.01, 0.05, 0.05, 0.10)$ \\
    \multicolumn{2}{c|}{}             & $R_\Delta$ & $\text{diag}(0.05, 0.10, 0.10, 0.30)$ \\
    \bottomrule
  \end{tabular}
  \label{tab:mppi_parameters}
\end{table}

\section{Methodology} \label{section:methodology}

\subsection{Quadrotor Dynamics} \label{sec:quadrotor_dynamics}

We adopt the frame convention of~\cite{Mueller2025Dynamics}: ${B}$
is the body-fixed frame, with $\ub_3^{B}$ perpendicular to the propeller
plane, and ${E}$ the earth-fixed inertial frame, with $\ub_3^{E}$
opposite to gravity. The state is $\pos := \bm{s}_{BT} \in \real^3$
(position relative to a fixed earth point $T$), $\rot := \rot^{BE} \in
\SO(3)$ (attitude), $\vel := \bm{v}_{BE} \in \real^3$ (velocity), and
$\angvel := \bm{\omega}^{BE} \in \real^3$ (body rate), so
$\x = [\pos, \rot, \vel, \angvel]$. As in Euler's law $\mathbf{J}^{B}_{B}
\dot{\bm{\omega}}^{BE} + \bm{\omega}^{BE} \times \mathbf{J}^{B}_{B}
\bm{\omega}^{BE} = \bm{n}^{B}$, the MPPI controller commands a collective
thrust $c$ and desired body rate $\angvel_d$, $\u = [c, \angvel_d]$.

Since $\u$ cannot be realized instantaneously, we map every sampled input
through a feasibility projection following~\cite{Minark2024ModelPP}: the
desired rate is saturated to $\omega_{xy,\max}, \omega_{z,\max}$, the
corresponding torque $\bm{n}_d = \mathbf{J}^{B}_{B}\dot{\angvel}_d + \angvel
\times \mathbf{J}^{B}_{B}\angvel$ is computed with $\dot{\angvel}_d =
(\angvel_d - \angvel)/\dt$, and the single-rotor thrusts $\bm{c}_{P,d} =
\bm{M}^{-1}[c, \bm{n}_d]^\top$ are recovered from the mixer matrix $\bm{M}$
relating total thrust and torque to the four propeller thrusts (arm length $l_{\text{arm}}$, torque constant $\kappa$). Clipping each rotor thrust to
$[c_{P,\min}, c_{P,\max}]$ and mapping back through $\bm{M}$ gives the
feasible $c_{\text{clip}}, \bm{n}_{\text{clip}}$, from which
\begin{equation}
  \angvel_{\text{clip}} = \angvel + \dt \cdot (\mathbf{J}^{B}_{B})^{-1}\left(\bm{n}_{\text{clip}} - \angvel \times \mathbf{J}^{B}_{B}\angvel\right). \label{eq:feasible_rate}
\end{equation}
We denote this projection $\u_{\text{clip}} = \Pi(\x, \u) =
[c_{\text{clip}}, \angvel_{\text{clip}}]$; every sampled input is passed through $\Pi$
before propagation.

The state is then propagated with a semi-implicit Euler discretization of
Newton's and Euler's laws,
\begin{equation}
  \label{eq:drone_dynamics}
  \begin{aligned}
    \rot_{k+1} & = \rot_k \odot \text{Exp}\!\left(\angvel_{\text{clip}} \, \dt \right), \\
    \bm{a}_k    & = \mathbf{R}(\rot_{k+1}) \begin{bmatrix} 0 \\ 0 \\ c_{\text{clip}}/m^B \end{bmatrix} + \bm{g}^E, \\
    \vel_{k+1} & = \text{sat}_{v_{\max}}\!\left(\vel_k + \bm{a}_k \, \dt\right), \\
    \pos_{k+1} & = \pos_k + \vel_{k+1}\, \dt,
  \end{aligned}
\end{equation}
where $\text{Exp}(\cdot)$ integrates the attitude kinematics $\dot{\rot} =
\tfrac{1}{2}\rot \odot [0, \angvel]^\top$ exactly over $\dt$, $m^B$ is the
vehicle mass, $\bm{g}^E \approx (0,0,-9.81)\,\si{\meter\per\second\squared}$,
and $\text{sat}_{v_{\max}}(\cdot)$ clips speed to $v_{\max}$.

\subsection{Safety handling} \label{sec:feasibility}

By construction, MPPI enforces no hard safety constraint on its own. In fact, if two individually collision-free rollouts pass an obstacle on opposite sides, their weighted average can point straight through it, so the resulting nominal trajectory is not guaranteed to inherit the safety of the rollouts it is built from. To address this, we validate the trajectory at every planning cycle in two
stages before it is applied. First, we forward-simulate the newly averaged
nominal action sequence through the dynamics to obtain the corresponding
state sequence, and check every predicted state for collision against the
map, as described in \autoref{sec:cost}. Second, and independently, each
individual rollout $k$ is checked over its full horizon: rollout $k$ is
marked infeasible if any of its predicted states $\pos^k_j$, $j = 0,
\dots, H{-}1$, is found in collision. If the averaged trajectory fails
the first check, or if every rollout is marked infeasible by the second,
the controller discards the newly computed plan and instead commands the
drone to hover at the last predicted state that was previously verified
safe. At the next planning cycle, if the freshly computed MPPI plan passes
both checks, tracking resumes from it; otherwise the vehicle continues to
hover at that same last verified-safe state.

Beyond collision avoidance, we additionally require the committed
trajectory to terminate at zero velocity. Because nothing in the sampling and averaging procedure by itself guarantees this, costs are traded off against one another in the
exponential weighting, and a highly weighted rollout can still end with
nonzero residual velocity. We therefore enforce zero terminal velocity
through a dedicated braking-tail construction.

We split every rollout into two segments: a free segment, of length
$H_f = H - h_{\text{brake}}$, over which the control is sampled and
optimized, and a braking tail of length $h_{\text{brake}} \in \mathbb{N}$,
over which the input is replaced by a deterministic braking policy. Every
rollout is thus generated as
\begin{equation}
  \bm{u}_j^k =
  \begin{cases}
    \bm{u}^{\text{nom}}_j + \delta \bm{u}_j^k, & 0 \le j < H_f, \\[2pt]
    \pi_{\text{brake}}(\bm{x}_j^k), & H_f \le j < H,
  \end{cases}
  \label{eq:braking_tail}
\end{equation}
where the braking policy $\pi_{\text{brake}}$ commands a collective thrust
and attitude that decelerate the vehicle along its current velocity
direction and level it against gravity,
\begin{equation}
  \pi_{\text{brake}}(\x) = \Pi\!\left(\x, \begin{bmatrix} m\|\bm{a}_g\| \\ \angvel_{\text{level}}(\rot, \hat{\bm{a}}_g) \end{bmatrix}\right), \quad
  \bm{a}_g = \mathbf{g} - a_{\text{brake}}\frac{\vel}{\|\vel\|},
\end{equation}
with $a_{\text{brake}} \le \min(a_{\text{brake}}^{\max}, \|\vel\|/\dt)$ the
commanded braking deceleration and $\angvel_{\text{level}}$ a proportional
attitude-leveling rate that steers the body $z$-axis toward $\hat{\bm{a}}_g$.
Choosing the tail length long enough,
\begin{equation}
  h_{\text{brake}} \ge \left\lceil \frac{v_{\max}}{a_{\text{brake}}^{\max} \, \dt} \right\rceil
  + \left\lceil \frac{2\arctan\!\left(a_{\text{brake}}^{\max}/\|\mathbf{g}\|\right)}{\omega_{xy}^{\max}\,\dt} \right\rceil ,
  \label{eq:hbrake_bound}
\end{equation}
is sufficient for the braking policy to cancel translational velocity and
level the attitude against gravity: the first term bounds the time needed
to cancel velocity at the maximum braking deceleration, and the second the
time needed to level the attitude at the maximum body rate, so that a
rollout generated under \eqref{eq:braking_tail} can end at rest. This,
however, is only a sufficient condition on the nominal, unperturbed policy:
because MPPI is not an optimization problem but a sampling scheme, the
perturbations $\delta \bm{u}_j^k$ applied within the free segment, together
with disturbances and model mismatch along the rollout, mean that
\eqref{eq:hbrake_bound} does not by itself guarantee that any particular
sampled rollout ends at zero velocity, so in case the averaged trajectory does not terminate at zero velocity we command the hover as backup, as previously explained.

\subsection{Perception of Obstacles and Occlusion} \label{sec:perception}

At every sensor frame, the registered point cloud $\mathcal{P}_t \subset \real^3$ produced by the state estimator FAST-LIO \cite{Xu2020FASTLIOAF} is integrated into a probabilistic occupancy grid $\mathcal{M}$ maintained by ROG-Map~\cite{Ren2023ROGMapAE}. Each measurement updates the log-odds of the cell it terminates in, while the cells traversed by the corresponding ray are updated as free, so that every cell of $\mathcal{M}$ carries one of three labels: occupied, known-free, or unknown, the last denoting a cell no ray has yet resolved. The grid slides with the vehicle, which bounds memory and query time independently of the extent of the environment, and is closed from below and above by a virtual ground and ceiling that delimit the admissible workspace.

Because the vehicle is not a point, collision checking is performed on a second layer $\mathcal{M}_{\text{inf}}$ in which every occupied cell is dilated by $n_{\text{inf}} = \lceil r_{\text{veh}} / r_{\text{inf}} \rceil$ cells, with $r_{\text{inf}}$ the resolution of the inflated layer and $r_{\text{veh}}$ the radius of the sphere enclosing the airframe. A predicted state can then be tested as a point against $\mathcal{M}_{\text{inf}}$ rather than as a volume against $\mathcal{M}$. Since consecutive predicted positions may be several cells apart at the speeds we consider, the test is applied to the segment joining them: the segment $[\pos^k_{j-1}, \pos^k_j]$ is traversed at the resolution of the inflated layer, and the rollout is declared in collision as soon as one traversed cell is occupied or falls outside the local map. This yields the first safety term of the cost. 

\begin{figure}[!t]
  \centering
  \includegraphics[width=\columnwidth]{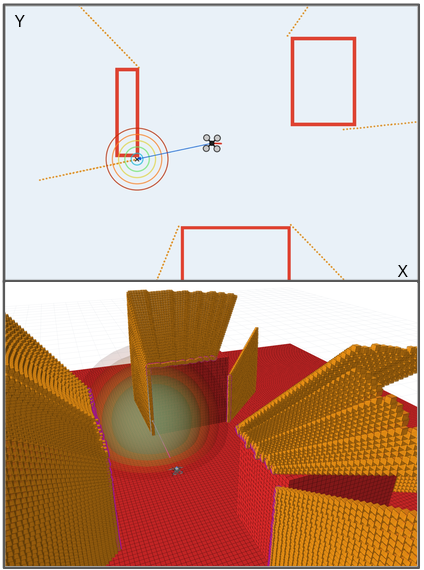}
    \caption{Occlusion boundary extraction for three obstacles surrounding the drone. \textit{Top:} top-down view; blue line point to the nearest
    occlusion voxel (red), with sphere color
    indicating the keep-out radius at increasing horizon times, from
    $t=0$\,s (blue) to $t=3$\,s (full horizon $H$, red). \textit{Bottom:} 3D
    view; yellow lines mark the occlusion lines of sight, with the
    expanding keep-out region (green concentric spheres) around the closest
    boundary point.}
  \label{fig:occlusion_boundaries_voxels}
\end{figure}
We extract, at every planning cycle, an occlusion boundary
$\mathcal{B} = \{\bm{b}_1, \dots, \bm{b}_M\} \subset \real^3$: a set of points
sampled along the silhouette edges of the locally observed occupancy map, i.e.,
the frontier between known-free and unknown space that is shadow-cast by an
occupied region. Physically, $\mathcal{B}$ marks the closest locations from which a hidden agent could emerge from occlusion.

Algorithm \autoref{alg:boundary} summarizes the extraction. The occupied cells of the
local map are first splatted into angular bins around the vehicle: each
cell is assigned to every bin whose angular extent it covers, as seen from the
current position, and each bin retains the smallest range among the cells
falling into it. The result is an ordered range profile of the observed
surfaces, which is the quantity a silhouette is defined on and which the raw
measurements do not provide in general. A pair of adjacent bins whose ranges
differ by more than a threshold $\tau$ then identifies a silhouette edge: the
nearer of the two bins corresponds to a surface the vehicle is looking past,
and the farther one to the shadow it casts. Two cases are distinguished: a
jump between two observed surfaces, and a surface adjacent to a bin containing
nothing at all, the last one corresponding to the end of a structure against
open space. The edge point is placed at the range of the nearer bin, on the
boundary between the two bins rather than at either bin center, which removes
a systematic half-bin bias. The gate is then emitted by sampling, at
map resolution, the segment of length $L$ that leaves the edge radially
outward, away from the vehicle. That segment is the surface a hidden agent
must cross to become visible, and the union of these segments over all edges
is $\mathcal{B}$.

\begin{algorithm}[t]
  \caption{3D occlusion boundary extraction}
  \label{alg:boundary}
  \begin{algorithmic}[1]
    \REQUIRE map $\mathcal{M}$, position $\pos$, range $R_{\text{sense}}$,
             resolutions $\Delta\theta, \Delta\phi$,
             elevation limits $[\phi_{\min}, \phi_{\max}]$,
             jump threshold $\tau$, gate length $L$, map resolution $r_m$
    \ENSURE occlusion boundary $\mathcal{B}$
    \STATE $\mathcal{O} \leftarrow \{\bm{o} : \|\bm{o} - \pos\|_2 \le R_{\text{sense}},\;
           \mathcal{M}(\bm{o}) = \text{occupied}\}$
    \STATE $d_{a,e} \leftarrow \infty$ for all azimuth/elevation bins $(a, e)$
    \FORALL{$\bm{o} \in \mathcal{O}$}
      \STATE $r \leftarrow \|\bm{o} - \pos\|_2$; \quad
             $(\theta, \phi) \leftarrow$ direction of $\bm{o} - \pos$
      \STATE $\alpha \leftarrow \arctan(r_{\text{vox}} / r)$
             \COMMENT{angular extent of the cell}
      \FORALL{bins $(a, e)$ covered by
              $[\theta \pm \alpha / \cos\phi] \times [\phi \pm \alpha]$
              within $[\phi_{\min}, \phi_{\max}]$}
        \STATE $d_{a,e} \leftarrow \min(d_{a,e}, r)$
      \ENDFOR
    \ENDFOR
    \STATE $\mathcal{B} \leftarrow \emptyset$
    \FORALL{neighbouring bins $i, j$ in azimuth or elevation}
      \IF{$d_i = d_j = \infty$ \OR $|d_i - d_j| < \tau$}
        \STATE \textbf{continue} \COMMENT{no surface, or continuous surface}
      \ENDIF
      \STATE $r_{\text{near}} \leftarrow \min(d_i, d_j)$; \quad
             $\bm{u} \leftarrow$ unit direction of the boundary between $i$ and $j$
      \STATE $\bm{e} \leftarrow \pos + r_{\text{near}} \, \bm{u}$
             \COMMENT{silhouette edge}
      \FOR{$s = 0$ \TO $L$ \textbf{step} $r_m$}
        \STATE $\mathcal{B} \leftarrow \mathcal{B} \cup \{\bm{e} + s \, \bm{u}\}$
      \ENDFOR
    \ENDFOR
    \RETURN $\mathcal{B}$
  \end{algorithmic}
\end{algorithm}

An advantage of operating on $\mathcal{M}$ rather than on the raw measurements
is that the method is sensor-agnostic. The splatting step of Algorithm \autoref{alg:boundary} reconstructs the ordered range profile from the map, so the same procedure applies across different sensing modalities and scanning patterns, including conventional spinning LiDARs (e.g., Velodyne VLP-32), non-repetitive scanning LiDARs such as the Livox MID-360S used in our setup, and depth cameras (e.g., Intel RealSense). The only requirement is that
the sensor delivers a point cloud that can be integrated into $\mathcal{M}$. A
second benefit follows from the same choice: because the map accumulates
measurements over time, the extracted boundary is not restricted to the
instantaneous field of view, and structures that have since left the field of view
continue to shadow the regions they occlude.

Given the online map, we obtain a distance field to the occlusion boundary, as
shown in \autoref{fig:occlusion_boundaries_voxels}, through a nearest-neighbor
query,
\begin{equation}
  d_{\mathcal{B}}(\pos) = \min_{\bm{b} \in \mathcal{B}} \left\| \pos - \bm{b} \right\|_2, \label{eq:boundary_distance}
\end{equation}
evaluated efficiently with a $k$-d tree built over $\mathcal{B}$ and rebuilt as
the boundary is re-extracted. When $\mathcal{B} = \emptyset$, i.e., no
occluding structure is currently observed, we set $d_{\mathcal{B}} \equiv
\infty$.

\subsection{Definition of the Cost} \label{sec:cost}

\begin{equation}
    \ell_{\text{goal}} = w_{\text{goal}} \, \|\pos^k_j - \pos^{\text{goal}}\|_2
    \label{eq:goal_cost}
\end{equation}
\begin{equation}
    \ell_{\text{goal, H-1}} = w_{\text{goal}}^{\text{term}} \, \|\pos^k_{H-1} - \pos^{\text{goal}}\|_2 \\
    \label{eq:terminal_goal_cost}
\end{equation}
\begin{equation}
  \ell_{\text{velocity}}(\pos^k_j, \vel^k_j) =
  \exp\!\left(-w_{\text{velocity}} \, \|\pos^k_j - \pos^{\text{goal}}\|_2^2\right) \cdot \|\vel^k_j\|_2^2,
  \label{eq:velocity_cost}
\end{equation}
\begin{equation}
    \ell_{\text{col}} = w_{\text{col}} \cdot \mathds{1}\!\left[\,
  [\pos^k_{j-1}, \pos^k_j] \cap \mathcal{M}_{\text{inf}} \neq \emptyset \,\right],
  \label{eq:collision_cost}
\end{equation}
\begin{equation}
  \ell_{\text{occ}} = w_{\text{col}} \cdot \mathds{1}\!\left[\, d_{\mathcal{B}}(\pos^k_j) < r_{\text{keep}}(t_j) \,\right], \label{eq:occlusion_cost}
\end{equation}
\begin{equation}
    \ell_{\text{control}} = \|\u^k_j\|^2_{R} \\
    + \|\Delta \u^k_j\|^2_{R_\Delta}
    \label{eq:control_cost}
\end{equation}
\begin{equation}
  r_{\text{keep}}(t) = d_{\text{safe}} + v_{\text{target}} \, t, \label{eq:keepout_radius}
\end{equation}

\begin{equation}
\begin{aligned}
  \S_k = \sum_{j=1}^{H-1} \Big[
      \ell_{\text{goal}}(\pos^k_j)
      + \ell_{\text{velocity}}(\pos^k_j, \vel^k_j) \\
      + \ell_{\text{col}}(\pos^k_{j-1}, \pos^k_j)
      + \ell_{\text{occ}}(\pos^k_j)
      + \ell_{\text{control}}(\u^k_j, \Delta \u^k_j)
  \Big] \\
  + \ell_{\text{goal, H-1}}(\pos^k_{H-1})
\end{aligned}
\label{eq:total_cost}
\end{equation}

Equations~\eqref{eq:goal_cost}--\eqref{eq:terminal_goal_cost} drive the rollout toward the goal, with the terminal term \eqref{eq:terminal_goal_cost} more heavily weighted so that reaching the goal at the end of the horizon is favored over merely approaching it along the way. Equation~\eqref{eq:velocity_cost} complements this with a softer, continuous shaping term: following the perception-aware weighting of \cite{Zhai_2026}, residual velocity is penalized in proportion to a Gaussian falloff in the squared distance to the goal, so the vehicle flies freely far from the goal but is increasingly discouraged from arriving at speed as it gets close. Equations~\eqref{eq:collision_cost} and~\eqref{eq:occlusion_cost} share the same hard-constraint weight $w_{\text{col}}$, since both encode conditions the plan should never realize rather than costs to be traded off continuously; \eqref{eq:collision_cost} rejects any rollout whose segment intersects a mapped obstacle, while \eqref{eq:occlusion_cost} rejects any rollout that enters the time-growing keep-out radius $r_{\text{keep}}(t_j)$ around the occlusion boundary defined in \eqref{eq:keepout_radius}, bounding exposure to a hidden agent that could have moved into the vehicle's path since the boundary was last observed. A rollout that ever triggers \eqref{eq:occlusion_cost} is thus treated as infeasible in the same sense as \eqref{eq:collision_cost}. Finally, Equation~\eqref{eq:control_cost} regularizes control effort and its rate of change, following the input and input-rate penalization used for agile MPPI flight in \cite{Minark2024ModelPP} with $R = \text{diag}(r_F, r_{\omega,xy}, r_{\omega,xy}, r_{\omega,z})$ and $R_\Delta = \text{diag}(r^\Delta_F, r^\Delta_{\omega,xy}, r^\Delta_{\omega,xy}, r^\Delta_{\omega,z})$ the input and input-rate weighting matrices.

\section{Results}

\subsection{Experimental Setup} \label{sec:setup}

To validate the effectiveness of our approach, we evaluate it both in
simulation and on real hardware. In simulation, a ray-casting sensor model
\cite{Kong2022MARSIMAL} attached to the vehicle body samples the scene, and
the resulting scan is consumed directly by ROG-Map~\cite{Ren2023ROGMapAE}.
In flight experiments, everything runs onboard a Jetson Orin Nano with
$8$\,GB of RAM. In both cases, rollouts are evaluated in parallel via OpenMP
across the available threads. Both the simulation tests and the real flight
experiments use the same MPPI parameters, listed in
\autoref{tab:mppi_parameters}.

\begin{figure}[!t]
  \centering
  \includegraphics[width=0.9\columnwidth]{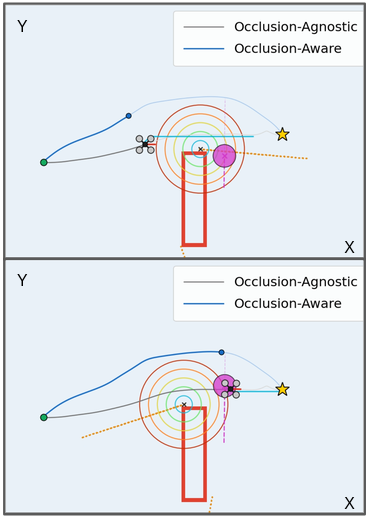}
  \caption{Simulation of an agent emerging from occlusion. The drone collides with the agent under baseline MPPI (gray), while OA-MPPI (blue) avoids the collision. The agent is shown as a purple sphere and speed $0.4$~m/s. Top: $t=2.8$~s; bottom: $t=3.9$~s.}
  \label{fig:sim_moving_agent}
\end{figure}

The ROG-Map occupancy grid covers $20 \times 20 \times 6$\,m at a
resolution of $r_m = 0.1$\,m. The inputs of Algorithm~\ref{alg:boundary}
are a query range of $R_{\text{sense}} = 4$\,m, an elevation window of
$[\phi_{\min}, \phi_{\max}] = [0^\circ, 40^\circ]$, a jump threshold of
$\tau = $~2\,m%TODO: value
, and a gate length of $L = 3$\,m, chosen to exceed the distance a
person at brisk walking speed could
cross within one replanning cycle. MPPI samples 500 rollouts over a
horizon of 30 steps ($\Delta t = \SI{0.1}{\second}$).

\begin{figure}[!t]
  \centering
  \includegraphics[width=0.9\columnwidth]{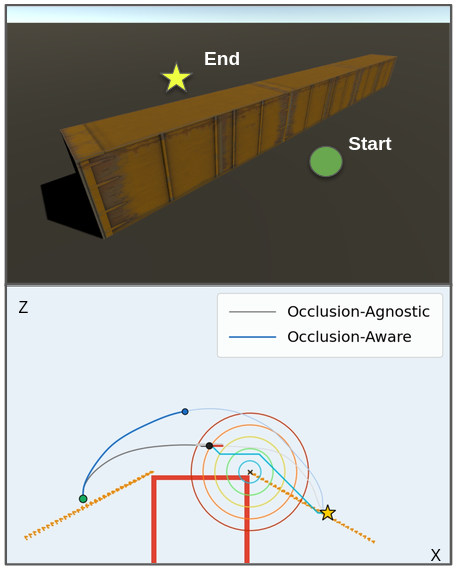}
  \caption{Top: simulation setup for flight over an occluding box. Bottom: side view in the $xz$-plane at $t=2.1$~s. OA-MPPI (blue) takes a higher path over the box than baseline MPPI (gray), increasing clearance from the occlusion boundary.}
  \label{fig:3d_occ}
\end{figure}

The MID-360 has a native vertical field of view spanning $-7^\circ$ to $52^\circ$ and is mounted with a $20^\circ$ forward tilt. The sensor provides point clouds at approximately \SI{10}{\hertz}. The range-jump threshold $\tau$ suppresses small range variations, including sensor noise, when identifying silhouette edges. The angular resolution $\Delta\theta$ is selected based on the point density at $R_{\text{sense}} = 4$~m, balancing spurious gaps from overly fine binning against merged boundaries from overly coarse binning.

\begin{figure*}[!t]
  \centering
  \includegraphics[width=\textwidth]{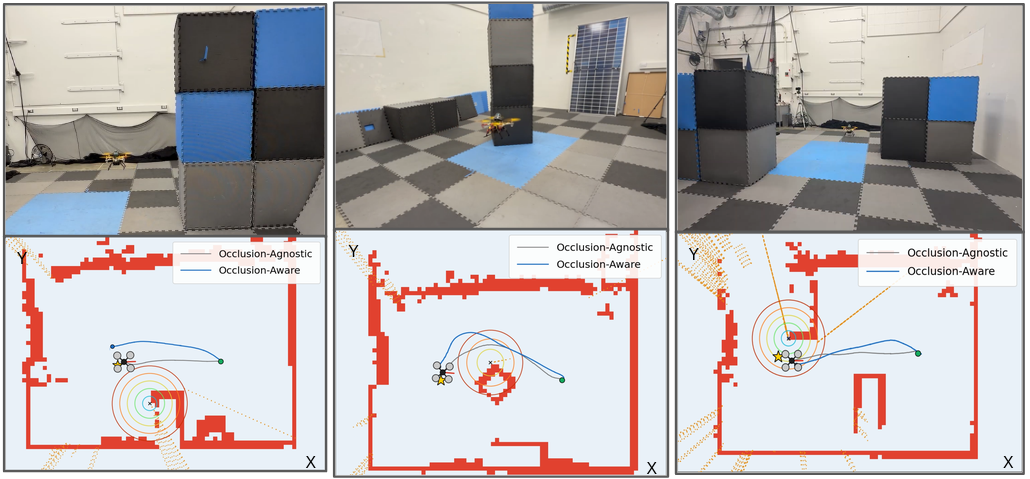}
  \caption{Experiments across the three test scenarios: single
  wall (left), pillar (middle), and two-wall corridor (right). \textit{Top:}
  physical setup, with the drone flying from its start position toward the
  goal near or behind the occluding structure(s). \textit{Bottom:} top-down
  occupancy map for the same run, showing the occlusion-agnostic (gray) and
  occlusion-aware (blue) trajectories from start (green dot) to goal (yellow star), with concentric contours marking the time-growing keep-out radius
  $r_{\text{keep}}(t)$ around the occlusion boundary and dashed orange lines
  indicating the occlusion lines of sight. The drone shape is only indicated on the occlusion-agnostic line for clarity. The goal is reached if within 0.5\,m from it.}
  \label{fig:real_flights}
\end{figure*}

\subsection{Simulation Tests}
\label{subsubsection:sim_test}

In the first simulation test shown in Fig.~\ref{fig:sim_moving_agent}, a moving agent emerges from occlusion and crosses the vehicle's path. This scenario is evaluated in simulation to avoid the risk of a physical collision. Baseline MPPI follows the subgoal selected along the A* path (light blue squared lines) and avoids the static obstacle, but passes too close to the occlusion boundary and collides with the agent (purple sphere). OA-MPPI steers around the expanding keep-out region, maintaining sufficient clearance to avoid the agent.

In the second test, shown in Fig.~\ref{fig:3d_occ}, we evaluate avoidance near a predominantly horizontal occlusion boundary, where a hidden agent could emerge vertically. The top image illustrates the scene in Unreal Engine, and the bottom plot shows the trajectories projected onto the $xz$-plane. With the occlusion cost defined in \autoref{sec:cost}, OA-MPPI takes a higher path over the obstacle than baseline MPPI, increasing clearance from the occlusion boundary. In both simulation tests, the keep-out region expands at $0.4$~m/s.

\subsection{Flight Experiments}
\label{subsubsection:experiments}

We conducted hardware flight experiments in the three scenarios shown in
\autoref{fig:real_flights}: a single wall with the goal positioned $3$~m
beyond it, a central pillar that the vehicle must navigate around, and a
two-wall corridor presenting two sources of occlusion, last two with goal $3.5$~m beyond. In each scenario,
we compared baseline MPPI and OA-MPPI using the same start and goal
positions. The keep-out expansion speed $v_{\text{target}}$ was set to
$0.4$~m/s for the single-wall and two-wall scenarios and $0.3$~m/s for
the pillar scenario.

\autoref{fig:real_flights} shows, for each scenario, the physical setup
alongside the corresponding top-down trajectory over the occupancy map. The
occlusion-aware trajectory (blue) consistently arcs further away from the
occlusion boundary than the baseline (gray), respecting the keep-out
constraint even in cases where the baseline trajectory cuts much closer to
it. \autoref{tab:real_flight_experiments} reports the corresponding time to
goal, distance traveled and mean velocity   for all three scenarios: the occlusion-aware
controller reaches the goal more slowly and along a longer path in every
case, reflecting the added detour needed to stay clear of the expanding
keep-out region rather than a failure to converge.

Following the configurations described in \ref{sec:setup},
and with rates measured from takeoff in the
experiments, the silhouette extractor keeps up with
the 10\,Hz registered point cloud, averaging $9.50$\,Hz in the
occlusion-aware runs ($9.57$\,Hz single wall, $9.61$\,Hz pillar and
$9.33$\,Hz two walls). The occlusion-agnostic planner averages
$9.59$\,Hz ($10.04$\,Hz single wall, $9.67$\,Hz pillar and
$9.07$\,Hz two walls). With the occlusion cost enabled, the
planner averages $8.47$\,Hz ($8.25$\,Hz single wall, $9.01$\,Hz pillar and $8.16$\,Hz two walls), about 12\% slower due to the overhead of
the boundary extractor and kd-tree queries against the nearest boundary
voxel at each rollout timestep. Despite this overhead, OA-MPPI remains fast enough for onboard replanning
in the tested flight scenarios, with only a modest reduction in update
rate relative to baseline MPPI.

\begin{table}[htbp]
  \footnotesize
  \centering
  \caption{Comparison of baseline MPPI and OA-MPPI across three flight scenarios.}
  \vspace{-0.3em}
  \setlength{\tabcolsep}{3pt}
  \begin{tabular}{l l c c c}
    \toprule
    Scenario & Method & Time [\si{\second}] & Dist. [\si{\meter}] & Vel. [\si{\meter\per\second}] \\
    \midrule
    \multirow{2}{*}{Single wall} & MPPI    & $2.49$ & $2.93$ & $1.18$ \\
                                  & OA-MPPI & $4.19$ & $3.76$ & $0.91$ \\
    \midrule
    \multirow{2}{*}{Pillar}      & MPPI    & $3.58$ & $3.77$ & $1.11$ \\
                                  & OA-MPPI & $3.74$ & $4.21$ & $1.14$ \\
    \midrule
    \multirow{2}{*}{Two walls}   & MPPI    & $3.06$ & $3.52$ & $1.15$ \\
                                  & OA-MPPI & $3.75$ & $3.85$ & $1.03$ \\
    \bottomrule
  \end{tabular}
  \label{tab:real_flight_experiments}
\end{table}

\section{Conclusion and Future Work} \label{sec:conclusion}

We presented OA-MPPI, a framework for UAV flight that extends MPPI control
with the ability to reason about occluded regions generated by obstacles in
the environment. The framework extracts occlusion boundaries directly
from the ego vehicle's local occupancy map, making the occlusion-detection
step agnostic to the specific onboard sensor. We penalize trajectories that enter an expanding keep-out region around the occlusion boundary within MPPI rollouts generated using nonlinear quadrotor dynamics. The region expands according to the assumed hidden-agent speed bound $v_{\text{target}}$, accounting for potential agent motion without assuming a specific trajectory. Simulation tests demonstrated increased clearance from occlusion boundaries and avoidance of an emerging agent that caused a collision under baseline MPPI. The clearance improvement over the baseline was also observed across three hardware flight scenarios, with the full perception and control pipeline running onboard the vehicle.

Three limitations motivate future work. First, the current method models hidden-agent motion using a fixed speed bound $v_{\text{target}}$. Future work could adapt this bound online based on context or replace the isotropic expansion with reachable sets tailored to the motion constraints of the expected agent type. Second, our presented hardware flight experiments did not include dynamic obstacles. Future work could incorporate dynamic obstacle detection and extend flight experiments to include both visible moving agents and agents emerging from occlusion. Third, expanding keep-out regions may restrict the available paths in narrow or heavily occluded environments, making the planner overly conservative. Future work could incorporate active perception to select viewpoints that reveal occluded space and reduce uncertainty about potential hidden agents.

\section*{Acknowledgment}
This work was supported by the Boeing Company.

\bibliographystyle{IEEEtran}
\bibliography{references}

\end{document}